\documentclass[a4paper,12pt,times,oneside,onecolumn,preprint]{elsarticle}
\usepackage[dvipsnames, svgnames, x11names]{xcolor}
 \usepackage{amsthm}
 \usepackage{graphics}
 \usepackage{array}
\usepackage{booktabs}
\usepackage{graphicx,subfigure}
\usepackage{float}
\usepackage[ruled,vlined,linesnumbered]{algorithm2e}
 \usepackage{graphicx}
 \usepackage{epsfig}
  \usepackage{paralist}
  \usepackage{epstopdf}
   \usepackage[colorlinks=true]{hyperref}

\usepackage{epsfig,graphicx,amssymb,amsmath}

\allowdisplaybreaks[4]
\newtheorem{theorem}{\textbf{Theorem}}

\usepackage{tikz}
\usetikzlibrary{arrows,automata}
\usepackage[latin1]{inputenc}
\usepackage{verbatim}
\usepackage{tikz}
\usepackage{tikz}
\usetikzlibrary{arrows,calc,shapes,decorations.pathreplacing}

\usepackage{colortbl}
\usepackage{color}
\usepackage{tabularx}
\usepackage{booktabs}
\usepackage{caption}
\usetikzlibrary{calc, positioning}
\pgfdeclarelayer{background}
\pgfdeclarelayer{foreground}
\pgfsetlayers{background,main,foreground}
\usepackage{ifthen}
\usepackage{animate}
\usepackage{graphicx,subfigure}
\usepackage{float}
\usepackage{epsfig,graphicx,amssymb,amsmath}
\usetikzlibrary{shapes.arrows}
\usepackage{textcomp,booktabs}

\usepackage{empheq}
\usepackage{amsfonts}
\usepackage{makecell}
\usepackage{amsmath}
\usepackage{epstopdf}
\usepackage{tikz}
\begin{document}
\begin{frontmatter}
\title{Research of Damped Newton Stochastic Gradient Descent Method for Neural Network Training}
\author[]{Jingcheng Zhou}
\author[]{Wei Wei}
\author[]{Zhiming Zheng}
\address{School of Mathematical Sciences, Beihang University}
\begin{abstract}
	First-order methods like stochastic gradient descent(SGD) are recently the popular optimization method to train deep neural networks (DNNs), but second-order methods are scarcely used because of the overpriced computing cost in getting the high-order information. In this paper, we propose the Damped Newton Stochastic Gradient Descent(DN-SGD) method and Stochastic Gradient Descent Damped Newton(SGD-DN) method to train DNNs for regression problems with Mean Square Error(MSE) and classification problems with Cross-Entropy Loss(CEL), which is inspired by a proved fact that the hessian matrix of last layer of DNNs is always semi-definite. Different from other second-order methods to estimate the hessian matrix of all parameters, our methods just accurately compute a small part of the parameters, which greatly reduces the computational cost and makes convergence of the learning process much faster and more accurate than SGD. Several numerical experiments on real datesets are performed to verify the effectiveness of our methods for regression and classification problems.
\end{abstract}

\begin{keyword}
	stochastic gradient descent,damped Newton
\end{keyword}
\end{frontmatter}

\section{Introduction}
First-order methods are popularly used to train deep neural networks(DNNs), such as stochastic gradient descent (SGD) and its variants that use momentum and acceleration and an adaptive learning rate\cite{ref2}. At each iteration, SGD calculates the gradient only on a small batch instead of the whole training data. Such randomness introduced by sampling the small batch can lead to better generalization of the DNNs\cite{ref3}. Recently there is a lot of work attempting to come up with more efficient first-order methods beyond SGD\cite{ref4}.  First-order methods are easy to implement and only require moderate computational cost per iteration, however, the requirement of adjusting their hyper-parameters (such as learning rate) becomes a complex task, and they are often slow to flee from areas where the loss function's hessian matrix is ill-conditioned.

Second-order stochastic methods have also been proposed for training DNNs because they take
far fewer iterations. Especially, second-order stochastic methods have the ability to escape from regions where the hessian matrix of the loss function is ill-conditioned and provide adaptive learning rates. But, there exist a main defect that it is practically impossible to compute and invert a full hessian matrix due to the massive parameters of DNNs. Efforts to conquer this problem include hessian-free inexact Newton methods, stochastic L-BFGS methods, Gauss-Newton and natural gradient methods and diagonal scaling methods\cite{ref5}. Besides, at the basement of those methods, more algorithms are generated such as SMW-GN and SMW-NG which use the Sherman-Morrison-Woodbury formula to cut the cost of computing the inverse matrix\cite{ref8} and GGN that combining GN and NTK\cite{ref1}. There are also some variances of the Quasi-Newton methods for approximating the hessian matrix\cite{ref6}\cite{ref7}.
\section{Main results}\label{section 3}
\subsection{feed-forward neural networks}\label{sub1}
Consider a simple fully connected neural networks with $L+1$ layers. For each layer, the number of the nodes is $n_l, 0\le l\le L$. In the following expressions, we put the bias into the weights for simplicity. Given an input $x$, add a component for $x$ to make the input as $x^{(0)}=(1,x^T)^T$ for the first layer. For the $l$-th layer, add a component for $x^{(l-1)}$, so $x^{(l)}=\sigma^{(l)}(W^{(l)}x^{(l-1)})$, where  $W^{(l)}=(w^{(l)}_1,w^{(l)}_2,\cdots,w^{(l)}_{n_l})^T\in R^{n_{l}\times (n_{l-1}+1)},w^{(l)}_j=(w^{(l)}_{0j},w^{(l)}_{1j},\cdots,w^{(l)}_{n_{l-1}j})^T\in R^{n_{l-1}+1}$ which denote the parameters between the $(l-1)$-th layer and the $j$-th node of $l$-th layer, $\sigma^{(l)}:R\to R$ is a nonlinear activation function and when $\sigma^{(l)}$ is used on a vector, it means $\sigma^{(l)}$ operates on every component of the vector, the output $f(\theta,x)=\sigma^{(L)}(W^{(L)}x^{(L-1)})$ where $\theta=\left({\rm vec}((W^{(1)})^T),{\rm vec}((W^{(2)})^T),\cdots,{\rm vec}((W^{(L)})^T)\right)^T$ in which ${\rm vec}(W)$ vectorizes the matrix $W$ by concatenating its columns.

Given the training data $(\boldsymbol{X},\boldsymbol{Y})=\{(x_1,y_1),(x_2,y_2),\cdots,(x_m,y_m)\}$, for training the neural networks, the loss function for minimization is generally defined as
\begin{equation}	L(\theta)=\dfrac{1}{m}\sum_{i=1}^{m}L_i(\theta,\boldsymbol{X},\boldsymbol{Y})=\dfrac{1}{m}\sum_{i=1}^{m}l(f(\theta,x_i),y_i)
\end{equation}
where $l(f(\theta,x_i),y_i)$ is a loss function. For the learning tasks, the standard Mean Square
Error(MSE) is always used for the regression problems and the Cross-Entropy Loss(CEL) is used for the classification problems. Note $\theta \in R^n$ where $n=\sum_{i=1}^{L}n_l(n_{l-1}+1)$.
\subsection{property of the loss function}
\begin{theorem}
	Given a neural network like the one above using MSE for loss function of regression problems that the number of the nodes for $L$-th layer is one. The last layer the activation function is used as identical function for regression problems. Then the hessian matrix of $L(\theta)$ to the parameters of last layer is positive semi-definite. Especially, when $\sigma^{(L-1)}(x)=relu(x)=max(0,x)$, the hessian matrix of $L(\theta)$ to the parameters of last but one layer is also positive semi-definite.	
\end{theorem}
\begin{proof}
	For simplicity, consider single one input first. Given the input $x_i$, the loss function should be \[L(W^{(L)})=\left(W^{(L)}x_i^{(L-1)}-y_i\right)^2\] and its gradient vector as well as hessian matrix should be
	\begin{equation}
		\dfrac{\partial L(W^{(L)})}{\partial W^{(L)}}=2\left(W^{(L)}x_i^{(L-1)}-y_i\right)x_i^{(L-1)},
	\end{equation}
	\begin{equation}
	\dfrac{\partial^2L(W^{(L)})}{\partial (w^{(L)})^2}=2x_i^{(L-1)}(x_i^{(L-1)})^T=A.
	\end{equation}
With the knowledge of matrix\cite{ref9}, if $x_i^{(L-1)}$ isn't a null vector, it can easily be got that $A$ is a positive semi-definite matrix with rank one. As for multiple inputs, the hessian is the addition of several positive semi-definite matrixes, so the hessian matrix is also positive semi-definite.

If the activation function $\sigma^{(L-1)}(x)=max(0,x)$, we can get the loss function \[L(W^{(L-1)})=(W^{(L)}\sigma^{(L-1)}(W^{(L-1)}x_i^{(L-2)})-y_i)^2\]
and the gradient vector as well as hessian matrix
\begin{equation}
\dfrac{\partial L}{\partial w_j^{(L-1)}}=2(W^{(L)}\sigma^{(L-1)}(W^{(L-1)}x_i^{(L-2)})-y_i)W^{(L)}_j\sigma^{'}((w_j^{(L-1)})^Tx_i^{(L-2)})x_i^{(L-2)},
\end{equation}
\begin{equation}
	\dfrac{\partial^2 L}{\partial w_j^{(L-1)}\partial w_k^{(L-1)}}=2W^{(L)}_jW^{(L)}_k\sigma^{'}((w_j^{(L-1)})^Tx_i^{(L-2)})\sigma^{'}((w_k^{(L-1)})^Tx_i^{(L-2)})x_i^{(L-2)}(x_i^{(L-2)})^T,
\end{equation}
\begin{equation}
	\dfrac{\partial^2 L}{\partial (v^{L-1})^2}=B\bigotimes C,
\end{equation}
where $\bigotimes$denotes kronecker product and
\[v^{(L-1)}=((w_1^{(L-1)})^T,(w_2^{(L-1)})^T,\cdots,(w_{n_{L-1}}^{(L-1)})^T)^T,\]
\[B\in R^{n_{L-1}\times n_{L-1}}, B_{jk}=2W^{(L)}_jW^{(L)}_k\sigma^{'}((w_j^{(L-1)})^Tx_i^{(L-2)})\sigma^{'}((w_k^{(L-1)})^Tx_i^{(L-2)}),\]\[C=x_i^{(L-2)}(x_i^{(L-2)})^T,\]
Let $u_j=W_j^{(L)}\sigma^{'}((w_k^{(L-1)})^Tx_i^{(L-2)}), u=(u_1, u_2, \cdots, u_{n_{L-1}})^T$, we can get $B=2uu^T$. Thus, it's easy to see $B$ and $C$ are
positive semi-definite matrix. Because the kronecker product of positive semi-definite matrices is also positive semi-definite\cite{ref10}, the hessian matrix $B\bigotimes C$ is positive semi-definite. Hence, the hessian of the multiple inputs is also positive semi-definite.
\end{proof}
\begin{theorem}
	Given a neural network like the one above using CEL for loss function of classification problems. The last layer $softmax$ is used as activation function for classification problems. Then the hessian matrix of $L(\theta)$ to the parameters of last layer is positive semi-definite.
\end{theorem}
\begin{proof}
	For simplicity, consider single one input first. Given the input $x_i$, the loss function should be \[L=-\sum_{j=1}^{n_L}Y_jlnx^{(L)}_{ij},\]
	where $Y_j$ is 1 when $x_i$ is the $j$-th class, otherwise 0. $x^{(L)}_{ij}=\dfrac{e^{(w^{(L)}_j)^Tx_i^{(L-1)}}}{\sum_{j=1}^{n_L}e^{(w^{(L)}_j)^Tx_i^{(L-1)}}},$	
	so \[L=-\sum_{j=1}^{n_L}Y_j(w^{(L)}_j)^Tx_i^{(L-1)}+ln(D),\]
	where $D=\sum_{j=1}^{n_L}e^{(w^{(L)}_j)^Tx_i^{(L-1)}}.$
	
	Then compute the gradient vector as well as hessian matrix
	\begin{equation}
		\dfrac{\partial L}{\partial w_j^{(L)}}=\left(\dfrac{e^{(w^{(L)}_j)^Tx_i^{(L-1)}}}{D}-Y_j\right)x_i^{(L-1)},
	\end{equation}
		\begin{equation}
		\dfrac{\partial^2 L}{\partial w_j^{(L)}\partial w_k^{(L)}}=
		\begin{cases}
		\dfrac{e^{(w^{(L)}_j)^Tx_i^{(L-1)}}(D-e^{(w^{(L)}_j)^Tx_i^{(L-1)}})}{D^2}x_i^{(L-1)}(x_i^{(L-1)})^T,j=k,\\
		-\dfrac{e^{(w^{(L)}_j)^Tx_i^{(L-1)}}e^{(w^{(L)}_k)^Tx_i^{(L-1)}}}{D^2}x_i^{(L-1)}(x_i^{(L-1)})^T,j\ne k.
		\end{cases}
		\end{equation}
	Let $v^{(L)}=((w_1^{(L)})^T,(w_2^{(L)})^T,\cdots,(w_{n_{L}}^{(L)})^T)^T,$	
	then we have
	\begin{equation}
		\dfrac{\partial^2 L}{\partial(v^{(L)})^2}=E\bigotimes F,
	\end{equation}
where
\[F=\dfrac{1}{D^2}x_i^{(L-1)}(x_i^{(L-1)})^T,\]
\begin{equation}
	E\in R^{n_L\times n_L},
	E_{jk}=
	\begin{cases}
	e^{(w^{(L)}_j)^Tx_i^{(L-1)}}(D-e^{(w^{(L)}_j)^Tx_i^{(L-1)}}),j=k,\\
	-e^{(w^{(L)}_j)^Tx_i^{(L-1)}}e^{(w^{(L)}_k)^Tx_i^{(L-1)}},j\ne k.
	\end{cases}
\end{equation}
	For any vector $z=(z_1,z_2,\cdots,z_{n_L})^T\in R^{n_L},$
	\begin{equation}
	z^TEz=\sum_{jk}^{n_L}z_jz_kE_{jk}=\dfrac{1}{2}\sum_{jk}^{n_L}e^{(w^{(L)}_j)^Tx_i^{(L-1)}}e^{(w^{(L)}_k)^Tx_i^{(L-1)}}(z_j-z_k)^2\ge 0.
	\end{equation}
Thus, $E$ and $F$ are positive semi-definite matrixes. Because the kronecker product of positive semi-definite matrices is also positive semi-definite\cite{ref10}, the hessian matrix $E\bigotimes F$ is positive semi-definite and the hessian matrix of the condition of multiple inputs is also positive semi-definite.	
\end{proof}

\section{Our innovation:Damped Newton Stochastic Gradient Descent}
As for the property of the loss function, the convexity of the loss function can be got for the last layer of DNNs. Thus, we can use Damped Newton method for parameters of the last layer and SGD for other layers. For each iteration, if the Damped Newton method is used first, then SGD, it is called Damped Newton Stochastic Gradient Descent(QN-SGD). Else, the SGD is used first, then Damped Newton method, which is called Stochastic Gradient Descent Damped Newton method(SGD-QN).

In traditional damped Newton method, adding scalar matrix is used to make the hessian matrix positive definite. The iteration is $\theta^{(t+1)}=\theta^{(t)}-(H^{t}+\lambda I)^{-1}g^{(t)}$. This method is hard to put into practise for the uncertain of $\lambda$. And the Quasi-Newton like others is to approximate the hessian matrix for cutting the computing cost, but all these methods consider the whole hessian matrix and lose much information to approximate.

Our optimization method is based on more elaborate analysis for the hessian matrix, in which the hessian matrix is not used entirely for accelerating the training. We just use the last layer's hessian matrix by damped Newton method. The iteration is $\theta^{(t+1)}=\theta^{(t)}-(H^{(t)}+H^{(t)}_{max}+\alpha I)^{-1}g^{(t)}$ where $H_{max}^{(t)}$ is the maximum value of the element of the $H^{(t)}$ and $\alpha$ is the damping coefficient. $H_{max}^{(t)}$ is to make the matrix positive definite and lower the conditional number at the same time. $\alpha$ is just to keep off the condition that $H^{(t)}$ is too bad to be a null matrix or very close to a null matrix like results on {\bf Shot Selection} \ref{fig:classification1}.

Our method combines the advantages of the stochastic gradient descent and damped Newton. On one hand, the main part of the parameters of DNNs using SGD method for training guarantees that the computing cost changes little. On the other hand, the parameters of the last layer using QN-SGN or SGD-QN for training can make the learning process converge more quickly with just a little more computing cost. When the training data is huge, the less number of iterations can cut much time for training the DNNs and overcome the cost of computing the second-order information of the last layer.
\section{Algorithm}
In this section, the QN-SGD and SGD-QN methods are summarized as follows.
\\\\
\begin{algorithm}[H]
	\caption{QN-SGD}
	\label{alg1}
\KwIn{Training Set $S_M=\{(x_1,y_1),(x_2,y_2),\cdots,(x_M,y_M)\}$, hyper-parameter $\alpha$, learning rate $\lambda$, batchsize $N$}
	Initialize the network parameter $\theta^{(0)}, t=0$
	
	{\bf for} $t=0,1,2,\cdots${\bf do}
	
	\ \ \ \ Randomly select a mini-batch $S_N\subset S_M$ of size $N$
	
	\ \ \ \ Compute $H_{last}^{(t)},g_{last}^{(t)}$ for the last layer and $H_{max}^{(t)}=\max(H_{last}^{(t)})$
	
	\ \ \ \ Compute $p^{(t)}=(H_{last}^{(t)}+(\alpha+H_{max}^{(t)})I)^{-1}$
	
	\ \ \ \ Set $\theta_{last}^{(t+1)}=\theta_{last}^{(t+1)}-p^{(t)}g_{last}^{(t)}$
	
	\ \ \ \ Compute $g_{front}^{(t)}$for other layers
	
	\ \ \ \ Set $\theta^{(t+1)}_{front}=\theta^{(t)}_{front}-\lambda\times g_{front}^{(t)}$
	
	{\bf end for}
\end{algorithm}

\bigskip
\begin{algorithm}[H]
	\caption{SGD-QN}
	\label{alg:algorithm1}
	\KwIn{Training Set $S_M=\{(x_1,y_1),(x_2,y_2),\cdots,(x_M,y_M)\}$, hyper-parameter $\alpha$, learning rate $\lambda$, batchsize $N$}
	Initialize the network parameter $\theta^{(0)}, t=0$
	
	{\bf for} $t=0,1,2,\cdots${\bf do}
	
	\ \ \ \ Randomly select a mini-batch $S_N\subset S_M$ of size $N$
	
	\ \ \ \ Compute $g_{front}^{(t)}$ for other layers
	
	\ \ \ \ Set $\theta^{(t+1)}_{front}=\theta^{(t)}_{front}-\lambda\times g_{front}^{(t)}$
	
	\ \ \ \ Compute $H_{last}^{(t)},g_{last}^{(t)}$ for the last layer and $H_{max}^{(t)}=\max(H_{last}^{(t)})$
	
	\ \ \ \ Computer $p^{(t)}=(H_{last}^{(t)}+(\alpha+H_{max}^{(t)})I)^{-1}$
	
	\ \ \ \ Set $\theta_{last}^{(t+1)}=\theta_{last}^{(t+1)}-p^{(t)}g_{last}^{(t)}$
	
	{\bf end for}
\end{algorithm}
\bigskip
The only difference of the QN-SGD and SGD-QN methods is that in each iteration, the damped Newton method or the stochastic gradient descent is used first.
\section{Numerical example}\label{section 4}
\subsection{Regression Problem}
\begin{figure}[H]
	\centering
	\includegraphics[scale=0.55]{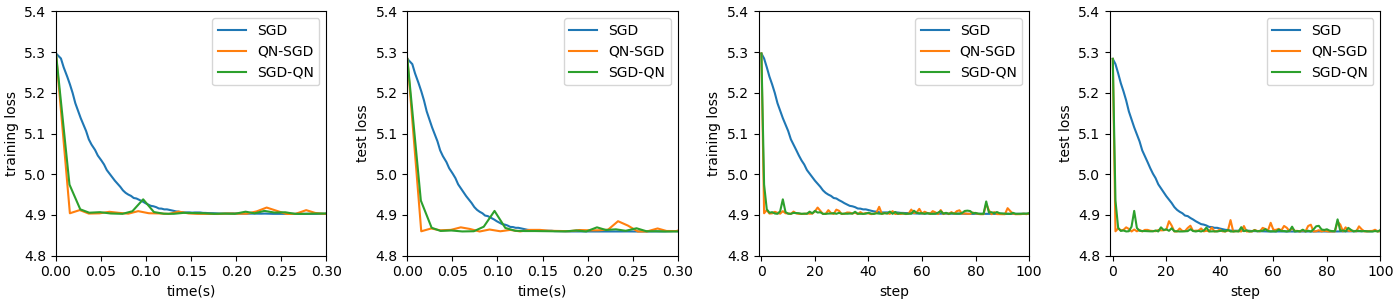}
	\caption{}
	\label{fig:regression1}
\end{figure}
Figure 1: Results on House Prices regression problem
with $N=50,\alpha=0,\lambda=0.01$
\\\\
\begin{figure}[H]
	\centering
	\includegraphics[scale=0.55]{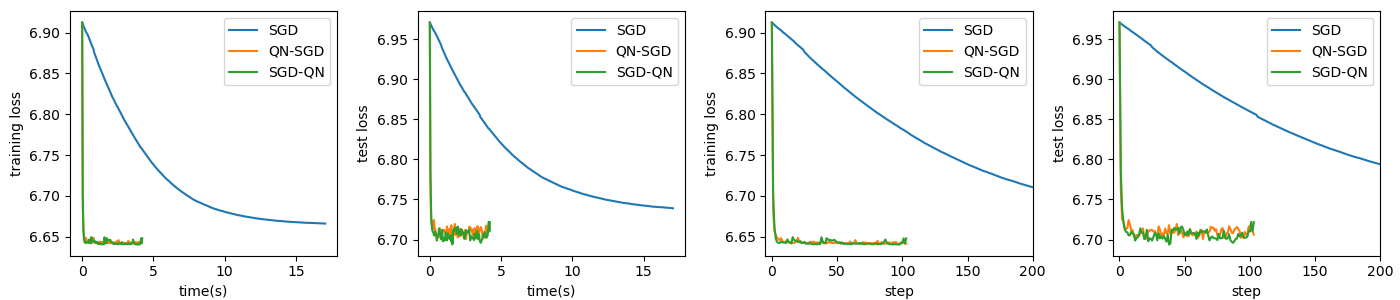}
	\caption{}
	\label{fig:regression2}
\end{figure}
Figure 2: Results on Housing Market regression problem
with $N=200,\alpha=0,\lambda=0.001$
\\\\
\begin{figure}[H]
	\centering
	\includegraphics[scale=0.55]{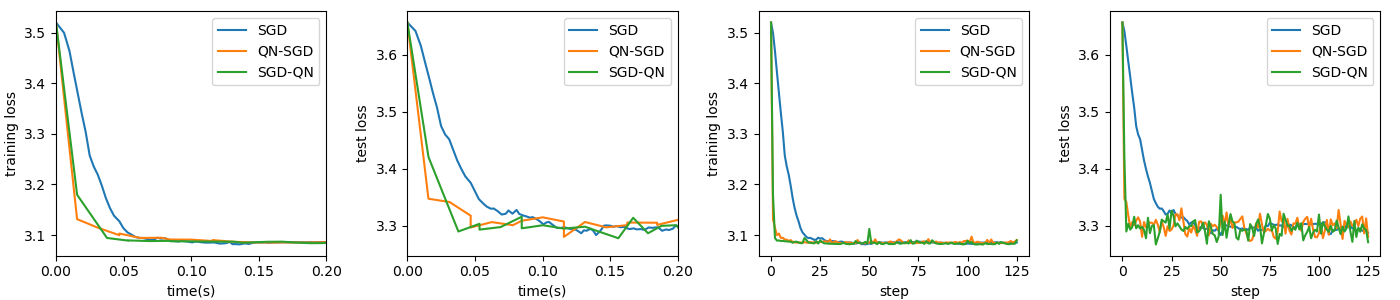}
	\caption{}
	\label{fig:regression3}
\end{figure}
Figure 3: Results on Cabbage Price regression problem
with $N=100,\alpha=0,\lambda=0.01$
\\\\
The performance of the algorithms QN-SGD and SGD-QN is compared with SGD, the experiments are performed on several regression problems, and both training loss and testing loss are provided. The data sets are scaled to have normal distribution.
\\\\
{\bf House Prices}: The training set is of size $1361$, and the test set is of size $100$. A neural network with one hidden layer of size 5 and sigmoid activation function is used, i.e., $(n_0,n_1,n_2)=(288,5,1)$, where the first and last layers are the size of input and output. The output layer has no activation function but with MSE, and the learning rate for SGD was set to be 0.01 the same as other methods for the purposes of comparison. Each algorithm is run for 20 epochs. The website is \url{https://www.kaggle.com/c/house-prices-advanced-regression-techniques/data}. The results are presented in Figure \ref{fig:regression1}.
\\\\
{\bf Housing Market}: The training set is of size $20473$, and the test set is of size $10000$. A neural network with one hidden layer of size 4 and sigmoid activation function is used, i.e., $(n_0,n_1,n_2)=(451,4,1)$, where the first and last layers are the size of input and output. The output layer has no activation function but with MSE, and the learning rate for SGD was set to be 0.001 the same as other methods for the purposes of comparison. Each algorithm is run for 5 epochs. The website is \url{https://www.kaggle.com/c/sberbank-russian-housing-market/data}. The results are presented in Figure \ref{fig:regression2}.
\\\\
{\bf Cabbage Price}: The training set is of size $2423$, and the test set is of size $500$. A neural network  with one hidden layer of size 6 and sigmoid activation function is used, i.e., $(n_0,n_1,n_2)=(4,6,1)$, where the first and last layers are the size of input and output. The output layer has no activation function but with MSE, and the learning rate for SGD was set to be 0.01 the same as other methods for purposes of comparison. Each algorithm is run for
5 epochs. The website is \url{https://www.kaggle.com/c/regression-cabbage-price/data}. The results are presented in Figure \ref{fig:regression3}.
\subsection{Classification Problem}
\begin{figure}[H]
	\centering
	\includegraphics[scale=0.55]{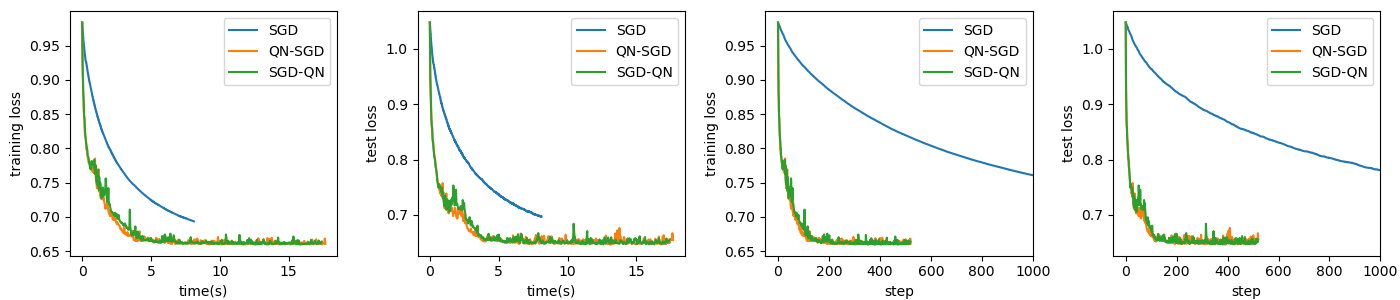}
	\caption{}
	\label{fig:classification1}
\end{figure}

Figure 4: Results on Shot Selection classification problem
with $N=200,\alpha=0.01,\lambda=0.01$

\begin{figure}[H]
	\centering
	\includegraphics[scale=0.55]{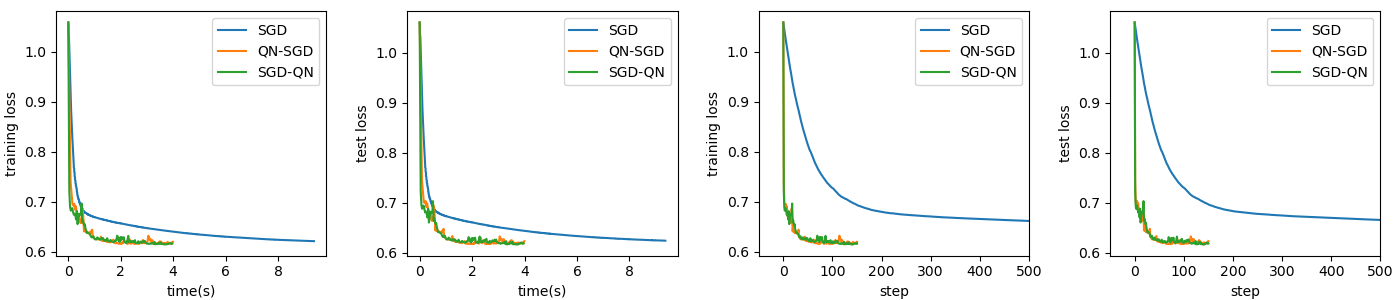}
	\caption{}
	\label{fig:classification2}
\end{figure}
Figure 5: Results on Categorical Feature Encoding classification problem
with $N=200,\alpha=0,\lambda=0.01$

\begin{figure}[H]
	\centering
	\includegraphics[scale=0.55]{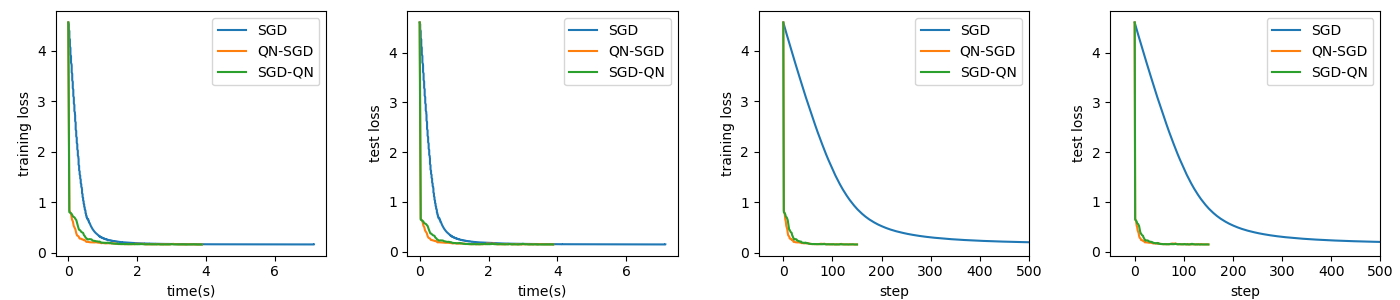}
	\caption{}
	\label{fig:classification3}
\end{figure}
Figure 6: Results on Categorical Feature Encoding classification problem
with $N=200,\alpha=0,\lambda=0.01$
\\\\
The performance of the algorithms QN-SGD and SGD-QN is compared with SGD, the experiments are performed on several classification problems, and both training loss and testing loss are provided. The data sets are scaled to have normal distribution.
\\\\
{\bf Shot Selection}: The training set is of size $20698$, and the test set is of size $10000$. A neural network with one hidden layer of size 6 and sigmoid activation function is used, i.e., $(n_0,n_1,n_2)=(228,6,2)$, where the first and last layers are the size of input and output. The output layer has softmax with cross entropy, and the learning rate for SGD
was set to be 0.01 the same as other methods for the purposes of comparison. Each algorithm is run for
25 epochs. The website is \url{https://www.kaggle.com/c/kobe-bryant-shot-selection/data}. The results are presented in Figure \ref{fig:classification1}.
\\\\
{\bf Categorical Feature Encoding}: The training set is of size $30000$, and the test set is of size $10000$. A neural network with one hidden layer of size 6 and sigmoid activation function is used, i.e., $(n_0,n_1,n_2)=(96,4,2)$, where the first and last layers are the size of input and output. The output layer has softmax with cross entropy, and the learning rate for SGD
was set to be 0.01 the same as other methods for the purposes of comparison. Each algorithm is run for
20 epochs. The website is \url{https://www.kaggle.com/c/cat-in-the-dat/data}. The results are presented in Figure \ref{fig:classification2}.
\\\\
{\bf Categorical Feature Encoding}: The training set is of size $30000$, and the test set is of size $10000$. A neural network with one hidden layer of size 4 and sigmoid activation function is used, i.e., $(n_0,n_1,n_2)=(369,4,2)$, where the first and last layers are the size of input and output. The output layer has softmax with cross entropy, and the learning rate for SGD
was set to be 0.01 the same as other methods for the purposes of comparison. Each algorithm is run for
20 epochs. The website is \url{https://www.kaggle.com/c/santander-customer-satisfaction/data}. The results are presented in Figure \ref{fig:classification3}.
\section{ Discussion of Results}
From the experimental results, it can be seen that QN-SGD and SGD-QN are always faster than SGD in terms of both steps and time, which accords with the provided analysis. Besides, when the dataset is huge, the QN-SGD and SGD-QN can be more efficient because the advantage of quick descent can exceed the cost of computing the second-order information.
\section{Conclusion and Future Research Directions}\label{section 5}
In this paper, we propose the QN-SGD and SGD-QN methods that combining the stochastic gradient descent and damped Newton method, which perform better than SGD in several data sets for training neural networks. Those methods also have the potential for application to more sophisticated models such as CNNs, ResNet and even GNNs. A promising future research topic is the study of property of the hessian matrix for each hidden layer to generate more quicker methods.

\end{document}